\def\year{2020}\relax
\documentclass[letterpaper]{article} 
\usepackage{aaai20}  
\usepackage{times}  
\usepackage{helvet} 
\usepackage{courier}  
\usepackage[hyphens]{url}  
\usepackage{graphicx} 
\usepackage{graphicx}  
\usepackage{amsmath,amssymb,amsfonts}
\usepackage{mathtools,algorithm}
\usepackage{algpseudocode}
\usepackage{textcomp}
\usepackage{xspace}
\usepackage{booktabs}
\usepackage{subfigure}

\newcommand{\citet}[1]{\citeauthor{#1} \shortcite{#1}}

\title{Effective Data Augmentation with Multi-Domain Learning GANs}

\author{Shin'ya Yamaguchi,\textsuperscript{\rm 1} Sekitoshi Kanai,\textsuperscript{\rm 1,2} Takeharu Eda\textsuperscript{\rm 1} \\ 
\textsuperscript{\rm 1}NTT Software Innovation Center\\ 
\textsuperscript{\rm 2}Keio University\\
Tokyo, Japan\\
\{shinya.yamaguchi.mw, sekitoshi.kanai.fu, takeharu.eda.bx\}@hco.ntt.co.jp
}

\begin{document}
\maketitle
\begin{abstract}
    For deep learning applications, the massive data development (e.g., collecting, labeling), which is an essential process in building practical applications, still incurs seriously high costs.
    In this work, we propose an effective data augmentation method based on generative adversarial networks (GANs), called Domain Fusion. 
    Our key idea is to import the knowledge contained in an outer dataset to a target model by using a multi-domain learning GAN. 
    The multi-domain learning GAN simultaneously learns the outer and target dataset and generates new samples for the target tasks. 
    The simultaneous learning process makes GANs generate the target samples with high fidelity and variety. 
    As a result, we can obtain accurate models for the target tasks by using these generated samples even if we only have an extremely low volume target dataset. 
    We experimentally evaluate the advantages of Domain Fusion in image classification tasks on 3 target datasets: CIFAR-100, FGVC-Aircraft, and Indoor Scene Recognition.
    When trained on each target dataset reduced the samples to 5,000 images, Domain Fusion achieves better classification accuracy than the data augmentation using fine-tuned GANs.
    Furthermore, we show that Domain Fusion improves the quality of generated samples, and the improvements can contribute to higher accuracy.
\end{abstract}

\newcommand{\DF}{{Domain Fusion}\xspace}
\newcommand{\df}{{domain fusion}\xspace}

\newcommand{\DA}{{DA}\xspace}
\section{Introduction}
Deep learning models have demonstrated state-of-the-art performance in various tasks using high dimensional data such as computer vision \cite{real2018regularized_amoebanet}, speech recognition \cite{seq2seq_attention_interspeech18}, and natural language processing \cite{NIPS2017_7181_attention_is_all_you_need}.
These models achieve the high performance by optimizing their millions of parameters through the training on labeled data.
Since the models can easily overfit the small data due to the enormous parameters, the generalization performance tends to be in proportion to the size of labeled data.
In fact, \citet{Sun_ICCV17} experimentally showed that the test performance on vision tasks could be improved logarithmically with the labeled data size.
To obtain higher performance of deep models, we must develop as many labeled data as possible by collecting data and attaching labels.
However, developing the labeled data becomes one of the main obstacles in the entire deployment of deep models since it requires a lot of time and high costs.

One of the most common techniques to alleviate the costs of labeled data developments is {\em data augmentation (DA)}.
To improve the performance of the target task (e.g., classification or regression), \DA amplifies the variation of existing labeled data (target data) by adding small transformations (e.g., random expansion, flip, and rotation).
Since \DA improves the performance despite its simplicity and has no dependency on network architectures, it is widely applied to many applications \cite{Krizhevsky_NIPS12,Ko_PPK15}.
However, when we train target models on low-volume datasets, the improvements by \DA is limited because \DA is designed to transform an existing sample into a slightly modified sample.
In other words, \DA does not generate truly {\em unseen} data, which have information not included in the data to be transformed.
For example, in image recognition, DA is not able to transform running-horse images into sitting-horse images.
Therefore, the benefit of DA is limited when we only have low-volume datasets.

Several methods \cite{Tran_NIPS17,Zheng_ICCV17,calimeri2017_biomedical_data_augmentation,Zhu2018bmvc_CGAN_Augmentation,antoniou_ICLR18W_DAGAN} have been presented to overcome the limitation of DA by applying generative adversarial networks (GANs, \citet{Goodfellow_NIPS14}).
GANs generate various and realistic data samples by learning data distributions; they can generate unseen samples from the learned distributions.
The existing methods employ this ability and use the generated samples as additional input for the target task.
Although these GAN-based methods succeed at improving the target performance, they assume that there is a sufficient volume of data for training GANs.
In fact, in the case of low volume data, the generated samples have less fidelity and variety and can degrade the target performance \cite{wang_transferring_iccv18,shmelkov_ECCV18_howgoodismygan}.
This is because low volume data has insufficient knowledge, and thus, we need to utilize supplementary knowledge for training GANs.
To train GANs with low-volume target data, \citet{wang_transferring_iccv18} proposed Transferring GANs (TGANs) which incorporate a fine-tuning technique into GANs.
However, Wang et al. experimentally show TGANs do not improve the generating performance very well when we have only 1 K target dataset.

In this paper, we propose {\em Domain Fusion} (DF), which is an effective data augmentation technique exploiting GANs trained on a target and another dataset.
To generate helpful samples, DF incorporates knowledge from the {\em outer domain}, which is another domain from the target, into a GAN.
Specifically, we train GANs on target and outer datasets, simultaneously unlike TGAN.
After training GANs, we use the generated samples in the target domain for the target tasks.
In order to generate the target samples explicitly, we adopt conditional GANs that can produce the conditioned samples by assigning class labels.
As a result, DF transfers the helpful knowledge of the outer domain into generated target samples via the shared parameters of GANs.
We call this training method  {\em multi-domain training}, and the trained GANs {\em multi-domain learning GANs}.

Furthermore, to enhance the quality of the generated samples, we propose two improvement techniques for DF.
First, we introduce a metric to select an outer dataset that includes knowledge to generate more helpful target samples.
An appropriate outer dataset needs to be selected for the target domain since the performance of DF depends on the choice.
To this end, we develop a new metric based on Fr\'echet inception distance (FID, \citet{heusel_ttur_nips17}) and multi-scale structural similarity (MS-SSIM, \citet{wang_msssim}) that focuses on the relevance between the target and outer domain, and the diversity of the outer samples.
Second, when generating samples from a GAN, we apply filtering to remove extremely broken samples that could lead to negative effects on target models.
For this purpose, we use {\em discriminator rejection sampling} (DRS,\citet{azadi_drs_iclr19}), which uses the information from a discriminator of a GAN to omit the bad samples.
We extend the DRS algorithm for conditional GANs to generate high-quality class-conditional samples.
Applying these improvements, we can generate more helpful target samples.

Our experimental results demonstrate that the samples from our GANs in DF more improve the accuracy in a low data regime compared to TGANs.
Furthermore, we show that our GANs can produce higher quality samples than TGANs in terms of FID and Inception Score.
We also experimentally confirm the correlation between the quality of generated samples and the classification accuracy.
More importantly, we show that the classifiers trained by a combination of DF and conventional \DA outperform the ones trained by only using conventional \DA.

Our main contributions are as follows:
\begin{itemize}
    \item We propose a new data augmentation method using GANs called \DF, which transfers knowledge of the outer dataset into the target models by using a GAN trained on multi-domain via the shared parameters.
          We also propose a metric for outer dataset selection, and modified DRS for filtering generated samples. 
    \item We confirm that the correlations between the quality of generated samples and the target-task performances in our experiments on CIFAR-100, FGVC-Aircraft and Indoor Scene Recognition in low-volume data regime.
          These results support that \DF improve the target models because of the high quality generated samples.
\end{itemize}

\section{Background}\label{sec:background}
\subsection{Generative Adversarial Networks}
A generative adversarial network (GAN) is composed of a {\it generator} network $G_{\theta}(z)$, and a {\it discriminator} network $D_{\phi}(x)$ \cite{Goodfellow_NIPS14}.
The $G$ generates fake samples from random noise $z\sim p_z$ and the $D$ has a role to distinguish an observation $x$ whether $x$ comes from generator $G(z)$ or data distribution $p_{\rm data}$.
The objective functions for training a discriminator and a generator are respectively formalized as follows:
\footnotesize
\begin{eqnarray}
    \label{eq:adv_loss_d}
    L_{D} &=& - \mathbb{E}_{x \sim p_{\rm data}}\log{D_{\phi}(x)}    \nonumber \\
          & & - \mathbb{E}_{z \sim p_z}\log{(1-D_{\phi}(G_{\theta}(z)))}, \\
    \label{eq:adv_loss_g}
    L_{G} &=& - \mathbb{E}_{z \sim p_z}\log{D_{\phi}(G_{\theta}(z)).} 
\end{eqnarray}
\normalsize
Through a tandem training of $G$ and $D$, $D$ learns to maximize the probability of assigning the ``real" label into real examples, whereas $G$ learns to maximize the probability of failing the distinction by the $D$.
When $G$ and $D$ converge to equilibrium point, the generator network $G$ produces realistic samples as good representation of data distribution $p_{\rm data}$.

In \DF, we use conditional GANs (cGANs) \cite{odena_acgan,miyato_cgans_iclr18} that generate samples conditioned by class labels.
The objective functions are given by rewriting Eq.~(\ref{eq:adv_loss_d}) and (\ref{eq:adv_loss_g}):
\footnotesize
\begin{eqnarray}
    \label{eq:cgan_loss_d}
    L_{D} &=& - \mathbb{E}_{x \sim p_{\rm data}}\log{D_{\phi}(x,y)}    \nonumber \\
              & & - \mathbb{E}_{z \sim p_z}\log{(1-D_{\phi}(G_{\theta}(z,y),y))}, \\
    \label{eq:cgan_loss_g}
    L_{G} &=& - \mathbb{E}_{z \sim p_z}\log{D_{\phi}(G_{\theta}(z,y),y)}. 
\end{eqnarray}
\normalsize
While there are several formulations for cGANs, we adopt a projection based conditioning \cite{miyato_cgans_iclr18} as our implementation of cGAN.
This approach concatenates the embedded conditional vector to the feature vector of the generator and discriminator to learn the condition.

\subsection{Data Augmentation with GANs}\label{sec:dagan}
There are several studies applying GANs into data augmentation schemes.
\citet{calimeri2017_biomedical_data_augmentation} have proposed an approach simply applying generated samples as additional datasets for medical imaging tasks.
\citet{Zhu2018bmvc_CGAN_Augmentation} have shown an application using conditional GANs for augmenting plant images.
For re-identification tasks in computer vision, the study of \cite{Zheng_ICCV17} has presented a training method with unconditional generated samples. 
\citet{Tran_NIPS17} have presented a way to train classification models with GANs in semi-supervised fashion.
Similarly to our work, these studies leveraged generated samples from GANs as supplementary training data for target models.
This is an intuitive and flexible strategy because we can easily use the generated samples as augment dataset like conventional \DA.
However, in low volume data, these types of data augmentation suffers from the problem of insufficient training a GAN as described in the next section.
In fact, \citet{shmelkov_ECCV18_howgoodismygan} have shown that the generated samples from low-data trained GANs degrade the accuracy of classifiers.
Our approach can help these existing GAN-based methods to reduce the negative effects of this problem since it improves the quality of the generated samples in the case of low volume data.

\subsection{Training GANs with Low Data Volume}\label{sec:lessdata}
In a low volume training data regime, \citet{wang_transferring_iccv18} have shown a fine-tuning technique for training of GANs, called {\em Transferring GANs}.
The authors tried to initialize weights of a GAN by leveraging pretrained generators and discriminators with greater volume outer datasets such as ImageNet.
They investigated the effect of the target data size by the experiments where GANs were pretrained on the outer dataset (ImageNet) and then fine-tuned to the target dataset (LSUN Bedrooms).
Their results showed that fine-tuned GANs generate high-quality samples in the case of large target data (18.5 of FID by 1M samples),
but relatively low-quality samples in the case of less volume target data (93.4 of FID by 1K samples).
Since 1K of target samples still requires us much effort for developing dataset, training of GANs with low data volume is still challenging.

\section{Domain Fusion}\label{sec:domainfusion}
In this section, we present \DF using multi-domain learning GANs.
A multi-domain learning GAN is trained on the target dataset and outer dataset simultaneously.
The procedure of \DF consists of the following three steps; 
(a) selecting an outer dataset, (b) multi-domain training a GAN, (c) sampling target labeled samples from the trained GAN.
In the rest of this section, we describe each of the steps.

\subsection{Selecting Outer Dataset}
First, we select an outer dataset that has useful knowledge for the target domain.
In this paper, we denote a dataset $S$ composed of $X$ and $Y$, where $X$ is a set of data samples (e.g., images) and $Y$ is a set of labels.
If we have a target dataset $S_{\rm T}$, the outer dataset $S_{\rm O}$ is selected from the candidates $\{ S_{i} \}$ according to ${\cal M}(S_{\rm T}, S_{i})$ which is our outer dataset metric of $S_i$ for $S_{\rm T}$:
\footnotesize
\begin{equation}\label{eq:so}
\begin{aligned}
    S_{\rm O} = \{ (x,y)\,|\,x \in X_{i}, y \in Y_{i}, (X_i,Y_i) \in S_i, \\
    i={\arg\min}_{i}\,{\cal M}(S_{\rm T}, S_{i})\}.
\end{aligned}
\end{equation}
\normalsize
In fact, it is non-trivial what metrics we should choose for outer dataset selection.
We propose a metric that makes account both the relevance between the target and outer dataset, and the diversity of outer samples (see Improvements Section).

\subsection{Multi-Domain Training}
Next, we train a conditional GAN; discriminator $D(x,y)$ to minimize Eq.~(\ref{eq:cgan_loss_d}) and generator $G(z,y)$ to minimize Eq.~(\ref{eq:cgan_loss_g}) on both $S_{\rm T}$ and $S_{\rm O}$.
The objective functions of the multi-domain training are defined as follows:
\footnotesize
\begin{eqnarray}
    \label{eq:mdl_d}
    L_{D} &=& \alpha L_{D_{\rm T}} + (1-\alpha) L_{D_{\rm O}}, \\
    \label{eq:mdl_g}
    L_{G} &=& \alpha L_{G_{\rm T}} + (1-\alpha) L_{G_{\rm O}},
\end{eqnarray}
\normalsize
where,
\footnotesize
\begin{eqnarray}
    \label{eq:d_t}
    L_{D_{\rm T}} &=& - \mathbb{E}_{x_{\rm T} \sim p_{\rm target}}\log{D_{\phi}(x_{\rm T}, y_{\rm T})}  \nonumber \\ 
            & & - \mathbb{E}_{z \sim p_z}\log{(1-D_{\phi}(G_{\theta}(z,y_{\rm T}),y_{\rm T}))}, \\
    \label{eq:d_o}
    L_{D_{\rm O}} &=& - \mathbb{E}_{x_{\rm O} \sim p_{\rm outer}}\log{D_{\phi}(x_{\rm O},y_{\rm O})}    \nonumber \\
            & & - \mathbb{E}_{z \sim p_z}\log{(1-D_{\phi}(G_{\theta}(z,y_{\rm O}),y_{\rm O}))}, \\
    \label{eq:g_t}
    L_{G_{\rm T}} &=& - \mathbb{E}_{z \sim p_z}\log{D_{\phi}(G_{\theta}(z,y_{\rm T}),y_{\rm T})}, \\
    \label{eq:g_o}
    L_{G_{\rm O}} &=& - \mathbb{E}_{z \sim p_z}\log{D_{\phi}(G_{\theta}(z,y_{\rm O}),y_{\rm O})},
\end{eqnarray}
\normalsize
and $0 \leq \alpha \leq 1$ is a hyperparameter balancing the learning scale between the target and outer dataset ($\alpha=0.5$ in default setting).
In each step of the optimization, we sample data from the both target and outer dataset, and then compute the objective functions.
For both the target and outer domain, we adopt conditional GANs (CGANs) because the labels allow GANs to generate the target samples explicitly.
Furthermore, GANs with labels can achieve a higher generation performance than one without the labels~\cite{ICML19_S3GAN_lucic}.
We assume that $Y_{\rm T}$ and $Y_{\rm O}$ are disjoint each other.
In the training, we can summarize $Y_{\rm O}$ into one class since the target tasks do not use labels of the outer dataset.
However, we experimentally found that class-wise training with $Y_{\rm O}$ as well as $Y_{\rm T}$ contributes to the higher quality of generated samples.
We infer that this is because $Y_{\rm O}$ makes the learning of the outer domain be easier, and such learned representations help to generate target samples.
The overall procedure of the multi-domain training is illustrated in Algorithm~\ref{alg:domain_fusion}.

\algrenewcommand{\algorithmicrequire}{\textbf{Input:}}
\algrenewcommand{\algorithmicensure}{\textbf{Output:}}
\begin{algorithm}[t]
    \caption{Multi-Domain Training of Domain Fusion}
    \label{alg:domain_fusion} 
    \begin{algorithmic}[1]
      { \footnotesize
      \Require{Set of target data $X_{\rm T}$, set of outer data $X_{\rm O}$, set of target labels $Y_{\rm T}$, set of outer labels $Y_{\rm O}$, batchsize $B$, learning rate $\eta_\theta, \eta_\phi$, scaling factor $\alpha$}
      \Ensure Trained Generator $G_\theta$ 
      \State  Randomly initialize parameters $\theta$, $\phi$
      \While {not convergence}
        \For { $k$ steps }
            \State $\{x^i_T\}^{B}_{i=1},\{y^i_{\rm T}\}^{B}_{i=1} \leftarrow$ {\tt GetSample}{($X_{\rm T},Y_{\rm T},B$)}
            \State $\{z^i_{\rm T}\}^{B}_{i=1} \leftarrow$ {\tt GenNoise}{($B$)}
            \State $L_{D_{\rm T}} \leftarrow -\sum^{B}_{i} \log D_{\phi}(x^i_{\rm T},y^i_{\rm T})$ \Comment{Eq.(\ref{eq:d_t})}
            \State ~~~~~~~~~~~~~~~$-\sum^{B}_{i} \log(1 - D_{\phi}(G_\theta(z^i_{\rm T},y^i_{\rm T}),y^i_{\rm T}))$
            \State $\{x^i_{\rm O}\}^{B}_{i=1},\{y^i_O\}^{B}_{i=1} \leftarrow$ {\tt GetSample}{($X_{\rm O},Y_{\rm O},B$)}
            \State $\{z^i_{\rm O}\}^{B}_{i=1} \leftarrow$ {\tt GenNoise}{($B$)}
            \State $L_{D_{\rm O}} \leftarrow -\sum^{B}_{i} \log D_{\phi}(x^i_{\rm O},y^i_{\rm O})$ \Comment{Eq.(\ref{eq:d_o})}
            \State ~~~~~~~~~~~~~~~$-\sum^{B}_{i} \log(1 - D_{\phi}(G_\theta(z^i_{\rm O},y^i_{\rm O}),y^i_{\rm O}))$
            \State $\phi$ $\leftarrow$ $\phi-\eta_\phi\nabla_{\phi}(\alpha L_{D_{\rm T}} + (1-\alpha)L_{D_O})$ \Comment{Eq.(\ref{eq:mdl_d})}
        \EndFor
        \State $\{y^i_{\rm T}\}^{B}_{i=1} \leftarrow$ {\tt GetLabel}{($Y_{\rm T},B$)}
        \State $L_{G_{\rm T}} \leftarrow -\sum^{B}_{i} \log D_{\phi}(G_\theta(z^i_{\rm T},y^i_{\rm T}),y^i_{\rm T}))$ \Comment{Eq.(\ref{eq:g_t})}
        \State $\{y^i_O\}^{B}_{i=1} \leftarrow$ {\tt GetLabel}{($Y_{\rm O},B$)}
        \State $L_{G_{\rm O}} \leftarrow -\sum^{B}_{i} \log D_{\phi}(G_\theta(z^i_{\rm O},y^i_{\rm O}),y^i_{\rm O}))$ \Comment{Eq.(\ref{eq:g_o})}
        \State $\theta$ $\leftarrow$ $\theta - \eta_\theta\nabla_\theta(\alpha L_{G_{\rm T}} + (1-\alpha)L_{G_{\rm O}})$ \Comment{Eq.(\ref{eq:mdl_g})}
      \EndWhile
      }
    \end{algorithmic}
\end{algorithm}

\subsection{Sampling Target Examples}
After training, we generate a set of new data samples $X_{\rm gen}$ from the trained generator $G(z,y)$ as follows:
\footnotesize
\begin{eqnarray}
    X_{\rm gen} = \{ x \,|\, x = G(z,y), z \sim p_z, y \in Y_{\rm T} \}.
\end{eqnarray}
\normalsize
Note that the input label $y$ is an element of $Y_{\rm T}$ since the purpose of a \DF is to augment the target dataset $S_{\rm T}$.
We generate equal amount of samples for each label.

In general, trained conditional GANs generate samples by only using a generator $G$.
However, the generated samples can include poor quality samples that have been rejected by the discriminator at the training.
To obtain more high-quality samples, we apply discriminator rejection sampling (DRS, \citet{azadi_drs_iclr19}).
In the next section, we show our modified DRS algorithm for conditional sampling.

Finally, the generated $X_{\rm gen}$ is integrated into the target dataset $S_{\rm T}$.
\footnotesize
\begin{eqnarray}
    S_{\rm aug} &=& \{(x,y)\,|\,x \in X_{\rm aug}, y \in Y_{\rm aug} \}, \\
    X_{\rm aug} &=& X_{\rm T} \cup X_{\rm gen}, \\
    Y_{\rm aug} &=& Y_{\rm T}
\end{eqnarray}
\normalsize
We assume that generated data $X_{\rm gen}$ derived from the generator $G(z,y \in Y_{\rm T})$ have attribute consistency of the specified labels $y \in Y_{\rm T}$.
Thus, the augmented dataset $S_{\rm aug}$ is directly used as the input for the target model training as the alternative of the target dataset $S_{\rm T}$.

\section{Improvements}\label{sec:improvement}
\subsection{Outer Dataset Selection Metric}\label{sec:odscore}
In \DF, the choice of an outer dataset for the target is a dominant factor determining both the target model performance and the quality of generated samples.
In order to select a proper outer dataset, we focus on the {\em relevance} between the target and outer dataset, and the {\em diversity} of an outer dataset.

\subsubsection{Relevance Between the Target and Outer Dataset}
In the context of transfer learning, measuring the relevance between outer and target domain is widely used to avoid {\em negative transfer}, i.e., the target models could perform worse than the case of non transferring.
For GANs, \citet{wang_transferring_iccv18} attempt to select the outer dataset by measuring Fr\'echet inception distance (FID, \citet{heusel_ttur_nips17}) to the target dataset.
An FID between two datasets $X_{i}$ and $X_{j}$ is computed on features of ImageNet pretrained Inception Net:
\footnotesize
\begin{equation}\label{eq:fid}
    {\rm FID}(X_{i},X_{j}) = \|\mu_{i} - \mu_{j}\|^{2}_{2} + {\rm Tr}(\Sigma_{i}+\Sigma_{j}-2(\Sigma_{i}\Sigma_{j})^{\frac{1}{2}}),
\end{equation}
\normalsize
where $\mu_i$ and $\Sigma_i$ are the mean and covariance of the feature vectors of Inception Net for input $X_{i}$.
A lower FID means that $X_{i}$ and $X_{j}$ are highly related to each other.
Following Wang et al., we adopt FID as part of our metrics to measure the relevance of the target and outer dataset.
In our use, FID is a more preferable than other relevance metrics (e.g., general Wasserstein distance and maximum mean discrepancy) because there is no need to train additional feature extractors or kernel functions for each pair of datasets.

\subsubsection{Diversity of an Outer Dataset}
In~\cite{wang_transferring_iccv18}, they also reported the limitation of FID to predict actual quality of the generated samples from fine-tuned GANs.
This indicates that even if the outer dataset is highly relevant to the target, the outer dataset does not necessarily improve the quality of the generated target samples.
Thus, only using FID is insufficient for proper outer dataset selection.

In \DF, we propose a metric with an additional perspective of {\rm diversity} to select an outer dataset.
We assume that an outer dataset with diverse samples is preferable for the target sample generation because the more diverse samples can contain more useful and general information for target sample generations.
In order to select the dataset containing more diverse samples, we exploit multi-scale structural similarity (MS-SSIM, \citet{wang_msssim}).
MS-SSIM is an approach to assess structural similarity in multi-scale, and it is well accepted as an evaluation method for image compression tasks.
Recently, MS-SSIM is used for evaluating the diversity of generated samples by GANs~\citet{odena_acgan,miyato_cgans_iclr18}.
We apply MS-SSIM to assess the diversity of existing datasets for selecting more helpful outer datasets.
An MS-SSIM of two data samples $x_i$ and $x_j$ is defined as follows:
\footnotesize
\begin{equation}
    {\rm SSIM}(x_i,x_j)\!=\!l_{M}(x_i,x_j)^{\alpha_M}{\textstyle\prod_{m=1}^{M}}c_{m}(x_i,x_j)^{\beta_j}s_{m}(x_i,x_j)^{\gamma_m},
\end{equation}
\normalsize
where $l=\frac{2\mu_{x_i}\mu_{x_j}+C_1}{\mu_{x_i}^{2}+\mu_{x_j}^{2}+C_1}$,
$c=\frac{2\sigma_{x_i}\sigma_{x_j}+C_2}{\sigma_{x_i}^{2}+\sigma_{x_j}^{2}+C_2}$, 
$s=\frac{\sigma_{x_i x_j}+C_3}{\sigma_{x_i}\sigma_{x_j}+C_3}$, and $M$ denotes a scale number.
$l$ is computed only once at the maximum $M$, and $c, s$ are computed at all scales.
$\mu_{x_i}$ and $\sigma_{x_i}$ are the mean and standard deviation of $x_i$.
$\sigma_{x_i x_j}$ is the covariance of $x_i$ and $x_j$.
$\alpha$, $\beta,$ and $\gamma$ represent the hyperparameters, and $C_{1}$, $C_{2},$ and $C_{3}$ are small constants computed by the dynamic range of the pixel values and scalar constants.
The ranges of MS-SSIM is between 0 (high diversity) and 1 (low diversity), and ${\rm MS\mathchar`-SSIM}(x_i,x_i)=1$.

To evaluate the diversity of a dataset, we calculate the mean MS-SSIM for all the combinations of the samples in the dataset.
\footnotesize
\begin{equation}\label{eq:msssim}
    \overline{{\rm SSIM}}(X) = \frac{\sum_{x_{i} \in X}^{x_{i} \neq x_{j}}\sum_{x_{j} \in X}^{x_{j} \neq x_{i}}{\rm SSIM}(x_{i},x_{j})}{(|X|^{2}-|X|)},
\end{equation}
\normalsize
where $|X|$ denotes the size of $X$.
We consider that the mean MS-SSIM indicates the diversity of the dataset.

\subsubsection{Outer Dataset Metric ${\cal M}$} 
By combining FID and mean MS-SSIM, we compute an outer dataset metric ${\cal M}$ for a target dataset $X_{\rm T}$ and an outer dataset $X_{\rm O}$ as follows:
\footnotesize
\begin{equation}
    {\cal M}(X_{\rm T},X_{\rm O}) = {\rm FID}(X_{\rm T}, X_{\rm O})\cdot\overline{{\rm SSIM}}(X_{\rm O})
\end{equation}
\normalsize
A lower ${\cal M}$ indicates a more proper outer dataset.
We aim to select an outer dataset with both high relevance to a target dataset and high diversity within the samples.
This metric helps to pick such outer datasets according to the multiplication of FID and MS-SSIM representing the relevance and diversity, respectively.
The role of MS-SSIM (diversity), which is in $[0,1]$, is to weight FID (relevance), which is in $[0,+\infty]$.
In Experimental Results Section, we show that FID and MS-SSIM complementarily contribute to choosing an appropriate outer dataset in practice.

\subsection{Filtering by Modified DRS}\label{sec:drs}
In general, after training of GANs, we obtain the generated samples from GANs by only using the generator.
This is because we implicitly assume a successfully trained generator can always generate the samples fooling the discriminator with a probability of 1/2 \cite{Goodfellow_NIPS14}.
However, since this assumption does not hold in real world, the generator can produce broken samples that are easily detected by the discriminator as fake.
For data augmentation, we must avoid such broken samples.

In order to filter out broken samples, we adopt discriminator rejection sampling (DRS, \citet{azadi_drs_iclr19}) to \DF.
DRS is a rejection sampling method proposed for GANs, which computes an acceptance probability for each sample by using the density ratio from the discriminator.
Since DRS cuts off the broken samples according to the acceptance probability, sampling with DRS produces more high-quality samples than one with a generator alone.

Since the original paper of DRS has only shown the algorithm for unconditional sampling, 
we cannot directly apply the algorithm to \DF, which requires conditional sampling for the data augmentation.
Therefore, we modify the DRS algorithm for conditional sampling.
The modification is to compute the density ratio for each class label.
In the original DRS, one density ratio is estimated for a GAN without considering classes.
This may cause losing the diversity of samples of a specific class, because the sampling difficulty varies according to each class \cite{brock2018biggan}.
By estimating the class-wise density ratio, we aim to coordinate the acceptance probability for each class.
Applying this modification, we can obtain class conditional generated samples with high fidelity and variety.
(Our modified algorithm is shown in the supplemental materials.)

\begin{table}[t]
    \centering
        \caption{List of outer datasets. Each dataset size is the total number of the train and test size expect for Pascal-VOC. }
        \label{tb:outerdataset}
        \scalebox{0.75}{
        \begin{tabular}{lrr}\toprule
            Dataset & Classes& Size \\
            \midrule
            Oxford 102 Flowers \cite{Nilsback08_flowers} & 102 & 8,189\\
            Stanford Cars \cite{KrauseStarkDengFei-Fei_3DRR2013_cars} & 196 & 16,185\\
            Food-101 \cite{bossard14_Food101} & 101 & 101,000\\
            Describable Textures (DTD) \cite{cimpoi14describing_DTD} & 47 & 5,640\\
            LFW \cite{LFWTech} & 1 & 13,000 \\
            SVHN \cite{netzer2011reading_svhn} & 10 & 99,289\\
            Pascal-VOC 2012 Cls. \cite{Everingham15_PascalVOC} & 20 & 5,717\\
            \bottomrule
        \end{tabular}
    }
\end{table}

\newcommand{\tenexp}[1]{$\times 10^{#1}$}
\begin{table*}[t]
    \centering
        \caption{Performance comparison among data augmentation using GANs (top-1 and top5 classification accuracy (\%), FID, and IS). 
                TGAN and DF are the cases of applying Pascal-VOC as an outer dataset that marks the best score of our metric $\cal{M}$ for all targets.
                TGAN-Best denotes the best cases of TGAN approach when using another outer dataset that achieves the best accuracy.
                AVG represents average scores of 7 outer datasets.
                Note that, when we used 100\% volume of CIFAR-100, FGVC-Aircraft, and ISR (without generated images and any other data augmentations), the classifiers respectively achieved 61.71\%, 30.25\%, and 27.27\% test accuracy under these conditions.}
        \label{tb:cmp_gen}
        \scalebox{0.68}{
        \begin{tabular}{lcccccccccccc}
            \toprule
                                         &\multicolumn{4}{c}{CIFAR-100} &\multicolumn{4}{c}{FGVC-Aircraft}&\multicolumn{4}{c}{Indroor Scene Recognition} \\
            \cmidrule(lr){2-5}\cmidrule(lr){6-9}\cmidrule(lr){10-13}
                                         & Top-1 Acc. & Top-5 Acc. &             FID              & IS         & Top-1 Acc. & Top-5 Acc.               & FID              & IS & Top-1 Acc. & Top-5 Acc.               & FID              & IS  \\
            \midrule
            Without DA                   & 27.2$\pm$0.1 & 54.3$\pm$0.5 & $-$ & $-$            & 22.6$\pm$2.3 & 48.4$\pm$3.4 & $-$& $-$ &24.0$\pm$2.0 &52.0$\pm$0.7 & $-$ & $-$\\
            CGAN                         & 26.5$\pm$0.4 & 53.6$\pm$0.3 & 59.2$\pm$0.9 & 4.99$\pm$0.07 & 23.6$\pm$0.6 & 50.9$\pm$0.7 & 110.7$\pm$2.8& 3.41$\pm$0.05 & 25.7$\pm$0.7 & 52.6$\pm$0.7& 97.9$\pm$0.1& 3.48$\pm$0.02 \\
            TGAN                         & 25.9$\pm$0.5 & 52.1$\pm$0.6 & 60.9$\pm$5.9 & 5.20$\pm$0.20 & 24.1$\pm$0.4 & 51.0$\pm$0.7 & 109.0$\pm$3.6& 3.45$\pm$0.03 & 24.0$\pm$0.2 & 51.0$\pm$1.4& 104.1$\pm$5.3 & 3.49$\pm$0.07\\
            DF (ours)                    & {\bf 28.9}$\pm${\bf 0.5} & {\bf 56.2}$\pm${\bf 0.4} &  {\bf 53.5}$\pm${\bf 1.7 }& {\bf 5.32}$\pm${\bf 0.03} & {\bf 27.3}$\pm${\bf 0.9} & {\bf 55.4}$\pm${\bf 0.4}  & {\bf 97.9}$\pm${\bf 1.6}&{\bf 3.53}$\pm${\bf 0.15} & {\bf 26.1}$\pm${\bf 0.7} & {\bf 53.8}$\pm${\bf 0.9}  & {\bf 96.5}$\pm${\bf 4.0}&{\bf 3.61}$\pm${\bf 0.08} \\
            \midrule
            TGAN-Best                    & 28.2$\pm$0.5 & 55.7$\pm$0.2 & 54.5$\pm$5.2 & 5.16$\pm$0.03 & 26.2$\pm$0.3 & 52.9$\pm$0.3 & 109.5$\pm$2.0 & 3.47$\pm$0.03 & 25.7$\pm$1.0 & 54.9$\pm$1.2 & 97.8$\pm$5.8 & 3.50$\pm$0.03\\
            TGAN-AVG                     & 26.7$\pm$1.4 & 53.6$\pm$1.9 & 60.5$\pm$3.5 & 4.98$\pm$0.22& 23.8$\pm$3.4 & 49.8$\pm$4.6 & 113.4$\pm$8.5 & 3.42$\pm$0.04 & 23.4$\pm$1.5 & 51.0$\pm$2.3 & 111.8$\pm$12.8 & 3.38$\pm$0.18\\ 
            DF-AVG                       & 28.1$\pm$0.9 & 55.1$\pm$1.5 & 56.3$\pm$2.5 & 5.24$\pm$0.24& 25.2$\pm$1.5 & 52.3$\pm$1.8 & 105.8$\pm$15.2& 3.47$\pm$0.06 & 24.2$\pm$1.2 & 52.4$\pm$1.8 & 106.5$\pm$13.9 & 3.46$\pm$0.25\\
            \bottomrule
        \end{tabular}
        }
\end{table*}
\section{Experimental Results}\label{sec:experiment}
In this section, we show the evaluation of \DF (DF) on the image classification task using three datasets: CIFAR-100, FGVC-Aircraft, and Indoor Scene Recognition.
We compare our proposed DF with the conditional GAN (CGAN) and Transferring GAN (TGAN).

\subsection{Settings}
\subsubsection{Target Datasets}
The target task was the image classification on CIFAR-100~\cite{krizhevsky09_cifar10}, FGVC-Aircraft~\cite{maji13fine-grained_aircraft}, and Indoor Scene Recognition (ISR)~\cite{Quattoni_CVPR09_Recognizing_Indoor_Scene}.
We used CIFAR-100 instead of CIFAR-10 because CIFAR-100 can contribute to a more realistic evaluation with a larger number of labels and fewer samples per class.
These three datasets are characterized by samples with different features; CIFAR-100 is composed of the classes with various modes (vegetables, cars, furniture, etc.), 
FGVC-Aircraft includes only one mode (airplane) and has fine-grained classes that slightly differ each other,
and ISR is also constructed by one mode (indoor scenes) but has more diverse and rough-grained information than FGVC-Aircraft.
To evaluate the performance in low volume data setting, we reduced each training set of CIFAR-100 (50,000 images), FGVC-Aircraft (6,667 images), and ISR (5,360 images) to 5000 images, which are randomly sampled for each class.
Note that although the reductions for FGVC-Aircraft and ISR are relatively smaller than one of CIFAR-100, they originally have small absolute dataset volume per class; they are difficult to train the models even if we use full of the datasets.
We trained conditional GANs, and then, trained the classification model by using the generated samples as the additional dataset.
At the test step, we used the original test images (CIFAR-100: 10,000 images, FGVC-Aircraft: 3,333 images, ISR: 1,340 images) to accurately evaluate the trained models.

\subsubsection{Outer Datasets}
Table~\ref{tb:outerdataset} describes the list of the candidate for the outer dataset.
These are image datasets of various domain that are often used for the evaluation of computer vision tasks.
At training of DF and TGAN, we used train and test sets of these outer datasets except for Pascal-VOC.
We used only train set of Pascal-VOC for training because Pascal-VOC is employed for the reverse-side evaluation which flips the target and outer dataset each other
(The reverse-side evaluation is appeared in the supplemental materials).
For fair evaluation of the outer datasets, we randomly sampled 5,000 images from each dataset, and used the samples for training GANs.
We coordinated the number of samples to equal among classes.
Since these datasets contain various images of resolutions, we resized all of the images into 32$\times$32 by bilinear interpolation.

\subsubsection{Implementation Details}
{\bf GANs.}
We used ResNet-based SNGAN~\cite{miyato_SNGAN_iclr18,miyato_cgans_iclr18} for 32$\times$32 resolution images as the implementation of conditional GANs.
The model architecture was the same as \cite{miyato_cgans_iclr18}.
We trained a GAN for 50k iterations with a batch of 256 using Adam ($\beta_{1}=0$, $\beta_{2}=0.9$)~\cite{kingma_adam_iclr14}.
Following \cite{heusel_ttur_nips17}, the learning rate of generators and discriminators were $1.0 \times 10^{-4}$ and $4.0 \times 10^{-4}$, respectively.
We linearly shifted both the learning rates to 0.
Moreover, to fairly evaluate the models for each outer dataset, we incorporated early stopping with Inception Score (IS)~\cite{Salimans_NIPS16}.
The trigger of early stopping was set by estimated IS in each 1,000 iterations for 12,800 generated samples.
We stopped training when the consecutive drop count of IS reaches to 5.
In multi-domain training, we set $\alpha=0.5$ for all experiments.
In order to use filtering by DRS, we added additional sigmoid layers into the discriminator of the conditional SNGAN, and trained the additional layers for 10,000 steps for each class label (the learning rate was $1.0 \times 10^{-7}$).
For TGAN, we trained the conditional GANs on an outer dataset for 50k iterations with the early stopping, 
and then fine-tuned the pretrained GANs for a target dataset in the same setting.

{\bf Classifiers.}
The architecture for the target classifier was ResNet-18 for 224$\times$224~\cite{he_resnet} with Adam optimizer for 100 epochs, batches of size 512. 
We selected the batch size by grid search over {128, 256, 512, 1024} on all three target datasets to maximize the average accuracy across the datasets.
The hyperparameters for Adam were $\alpha_{\rm Adam} = 2.0\times 10^{-4}$, $\beta_{1}=0$, $\beta_{2}=0.9$.
We applied no conventional data augmentation (e.g., flip, rotation) to the input images without noted.
We used 50,000 samples (4,000 real images + 46,000 generated images) as training set, and 1,000 real images as validation set.
In all cases, we run the test for measuring mean accuracy on each test set of the target datasets.

\subsubsection{Evaluation Metrics}
We evaluated DF on the two aspects: the performance of target classification models and the quality of generated samples on target domain.
For the classifiers, we assessed the performance by top-1 and top-5 accuracy.
The sample quality was measured by Fr\'echet Inception Distance (FID)~\cite{heusel_ttur_nips17} and Inception Score (IS)~\cite{Salimans_NIPS16}.
For each target dataset, we computed FID and IS with 128 generated samples per class.
FID was calculated between the generated samples and the real images in the 100\% volume train set.
In all experiments, we trained GANs and classifiers three times, and show the mean and standard deviation of accuracy, FID, and IS.

\begin{table*}[t]
    \centering
        \caption{Ablation study of \DF}
        \label{tb:cmp_ablation}
        \scalebox{0.60}{
        \begin{tabular}{lcccccccccccc}
            \toprule
                                         &\multicolumn{4}{c}{CIFAR-100} &\multicolumn{4}{c}{FGVC-Aircraft} & \multicolumn{4}{c}{Indoor Scene Recognition}  \\
            \cmidrule(lr){2-5}\cmidrule(lr){6-9}\cmidrule(lr){10-13}
                                         & Top-1 Acc. & Top-5 Acc. &             FID              & IS         & Top-1 Acc. & Top-5 Acc.               & FID              & IS & Top-1 Acc. & Top-5 Acc.               & FID              & IS \\
            \midrule
            CGAN with DRS                                   & 27.3$\pm$0.3 & 54.5$\pm$1.3 & 58.7$\pm$0.8 & 5.05$\pm$0.01 & 24.6$\pm$0.8 & 52.4$\pm$0.9 & 110.0$\pm$3.5 & 3.42$\pm$0.09  & 24.8$\pm$0.9 &  52.8$\pm$0.5 & 99.9$\pm$6.8 & 3.42$\pm$0.06\\
            TGAN with DRS                                   & 26.6$\pm$1.5 &  53.5$\pm$1.2 & 59.9$\pm$5.9 & 5.22$\pm$0.03  & 24.4$\pm$1.2 & 52.2$\pm$0.6 & 107.4$\pm$3.0 & 3.49$\pm$0.05 & 24.9$\pm$0.8 & 53.2$\pm$1.4 & 103.9$\pm$2.5 & 3.44$\pm$0.09\\
            DF w/o $\cal{M}$ and DRS (Worst)                & 25.5$\pm$0.3 &  52.4$\pm$0.2 & 60.9$\pm$0.1 & 4.75$\pm$0.13  & 24.2$\pm$0.3 & 50.9$\pm$1.8 & 105.2$\pm$5.8 & 3.35$\pm$0.01 & 24.2$\pm$0.3 & 50.9$\pm$1.8 & 105.2$\pm$5.8 & 3.35$\pm$0.01 \\
            DF w/o DRS                                      & 28.3$\pm$0.7 & 55.7$\pm$0.5 & 54.9$\pm$2.4 & 5.16$\pm$0.04& 27.0$\pm$0.5 & 54.0$\pm$0.3 &  98.4$\pm$2.6 & 3.50$\pm$0.05 & 25.4$\pm$0.1 & 53.4$\pm$1.7 & 99.0$\pm$1.3 & 3.57$\pm$0.06\\
            DF                                              & {\bf 28.9}$\pm${\bf 0.5} & {\bf 56.2}$\pm${\bf 0.4} &  {\bf 53.5}$\pm${\bf 1.7 }& {\bf 5.32}$\pm${\bf 0.03} & {\bf 27.3}$\pm${\bf 0.9} & {\bf 55.4}$\pm${\bf 0.4}  & {\bf 97.9}$\pm${\bf 1.6}&{\bf 3.53}$\pm${\bf 0.15}& {\bf 26.1}$\pm${\bf 0.7} & {\bf 53.8}$\pm${\bf 0.9}  & {\bf 96.5}$\pm${\bf 4.0}&{\bf 3.61}$\pm${\bf 0.08}  \\
            \bottomrule
        \end{tabular}
        }
\end{table*}

\begin{table}[t]
    \centering
        \caption{Performance comparison to conventional \DA}
        \label{tb:cmp_da}
        \scalebox{0.60}{
        \begin{tabular}{lcccccc}\toprule
            & \multicolumn{2}{c}{CIFAR-100} & \multicolumn{2}{c}{FGVC-Aircraft} & \multicolumn{2}{c}{Indoor Scene Recognition} \\
            \cmidrule(lr){2-3}\cmidrule(lr){4-5} \cmidrule(lr){6-7}
            & Top-1 Acc. & Top-5 Acc. & Top-1 Acc. & Top-5 Acc. & Top-1 Acc. & Top-5 Acc.  \\
            \midrule
            cDA    &  30.7$\pm$0.7 & 57.3$\pm$0.3 &  29.6$\pm$0.9 & 58.5$\pm$1.6 & 31.0$\pm$0.3 & 59.6$\pm$0.7  \\
            DF+cDA &  {\bf 32.1}$\pm${\bf 0.7} & {\bf 59.2}$\pm${\bf 0.4} &  {\bf 31.2}$\pm${\bf 0.7} & {\bf 60.2}$\pm${\bf 1.0} &  {\bf 32.4}$\pm${\bf 1.7} & {\bf 61.6}$\pm${\bf 1.1} \\
            \bottomrule
        \end{tabular}
        }
\end{table}
\begin{figure*}[t]
    \centering
        \subfigure[CIFAR-100]{\includegraphics[width=5.5cm]{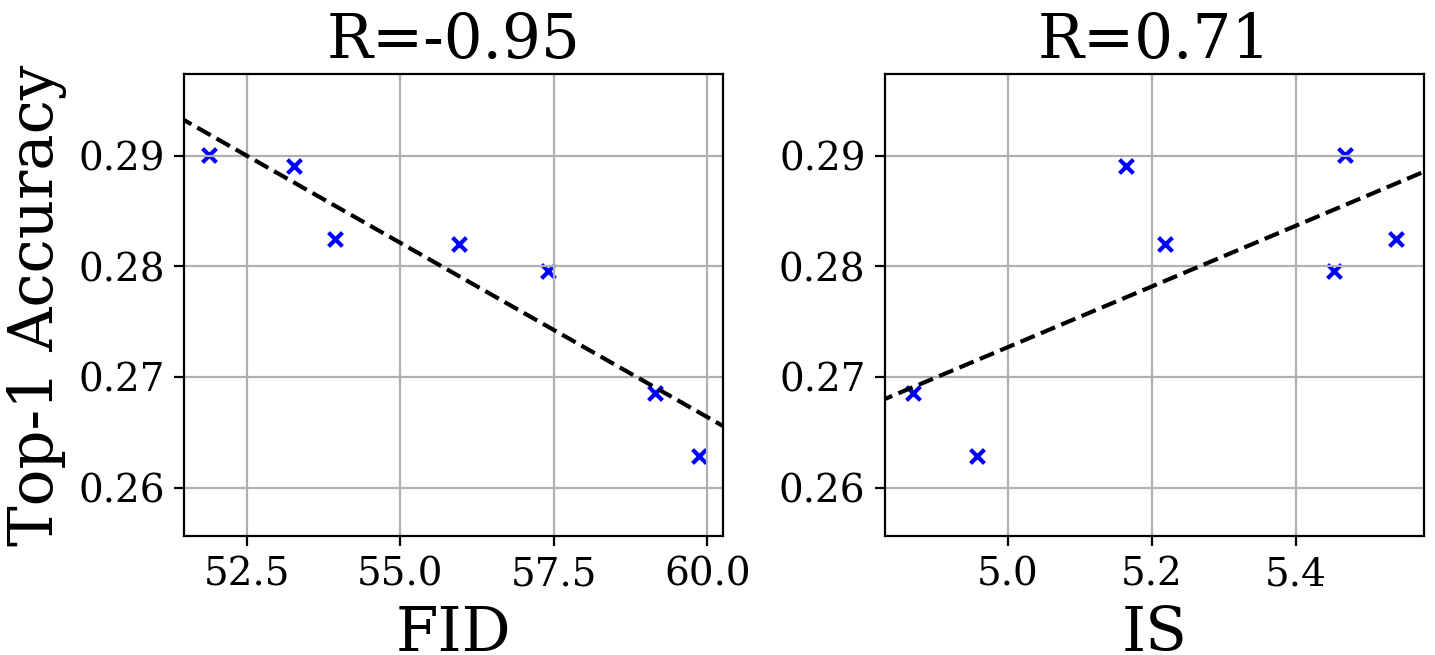}}
        \subfigure[FGVC-Aircraft]{\includegraphics[width=5.5cm]{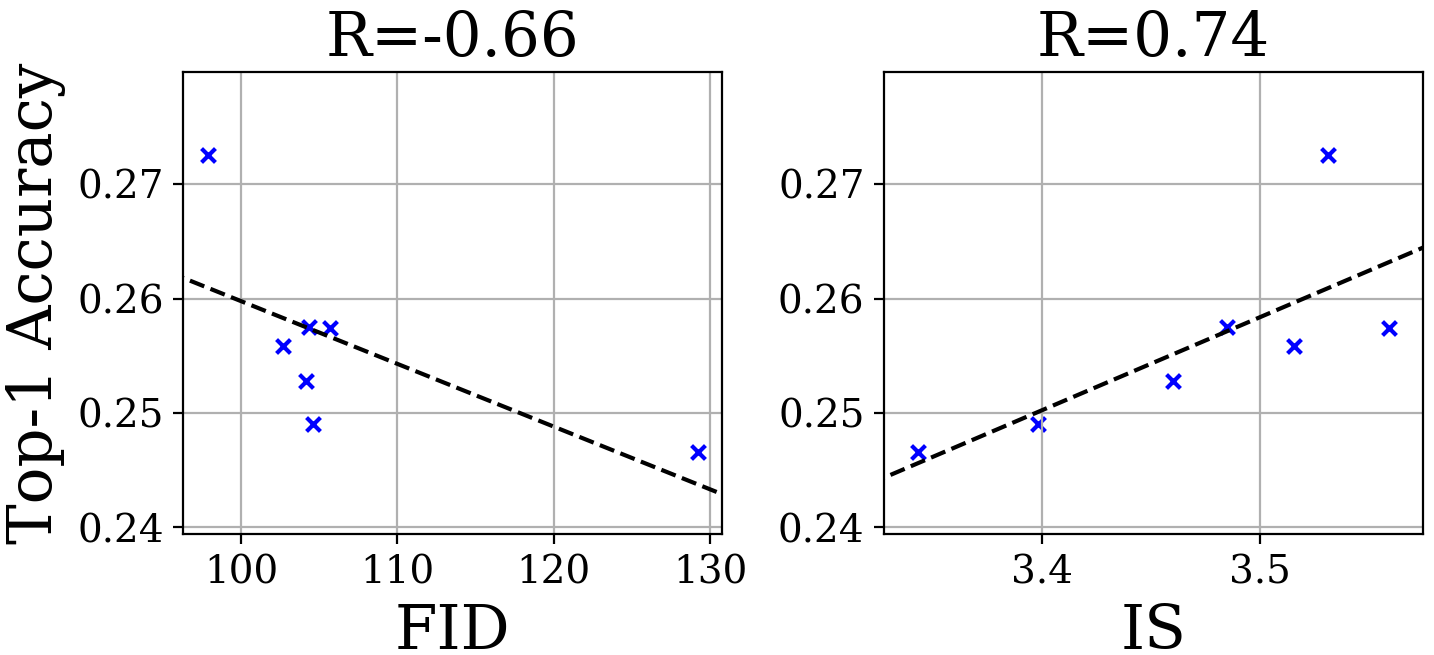}}
        \subfigure[Indoor Scene Recognition]{\includegraphics[width=5.5cm]{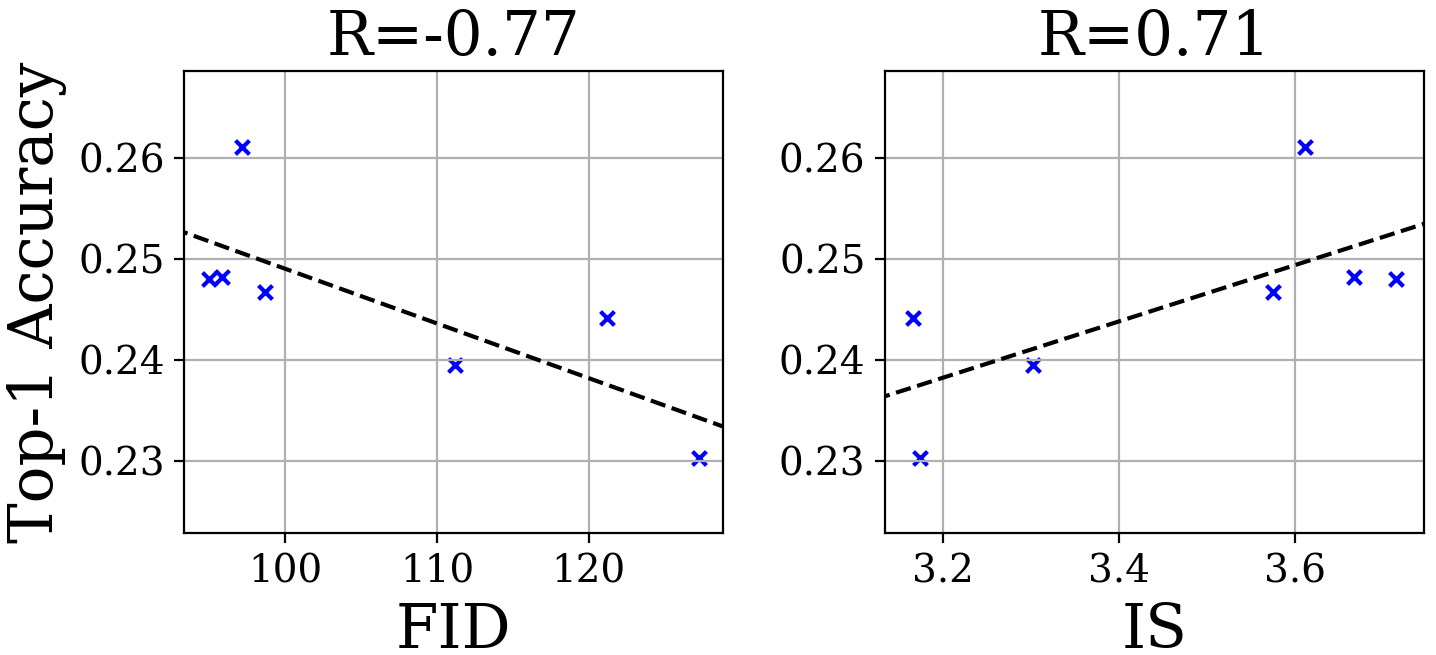}}
        \caption{Correlation between generated sample quality and top-1 accuracy }
        \label{fig:quality_vs_acc}
\end{figure*}
\begin{figure}[t]
    \centering
        \subfigure[CIFAR-100]{\label{fig:eval_m_cifar}\includegraphics[width=7.5cm]{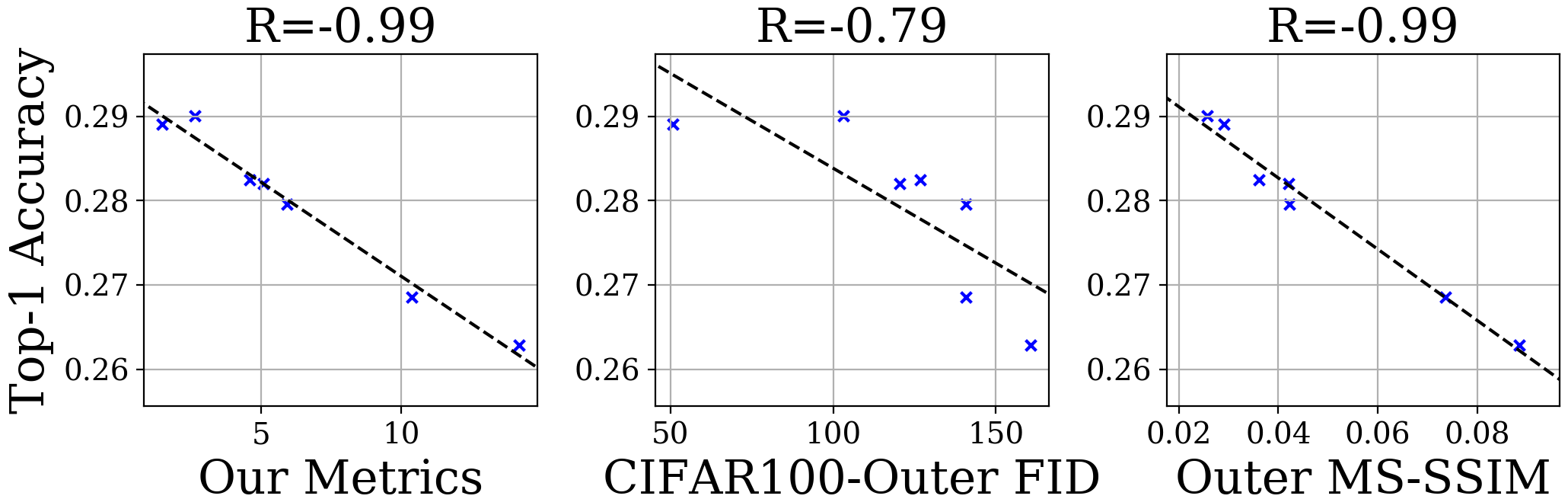}}
        \subfigure[FGVC-Aircraft]{\label{fig:eval_m_aircraft}\includegraphics[width=8cm]{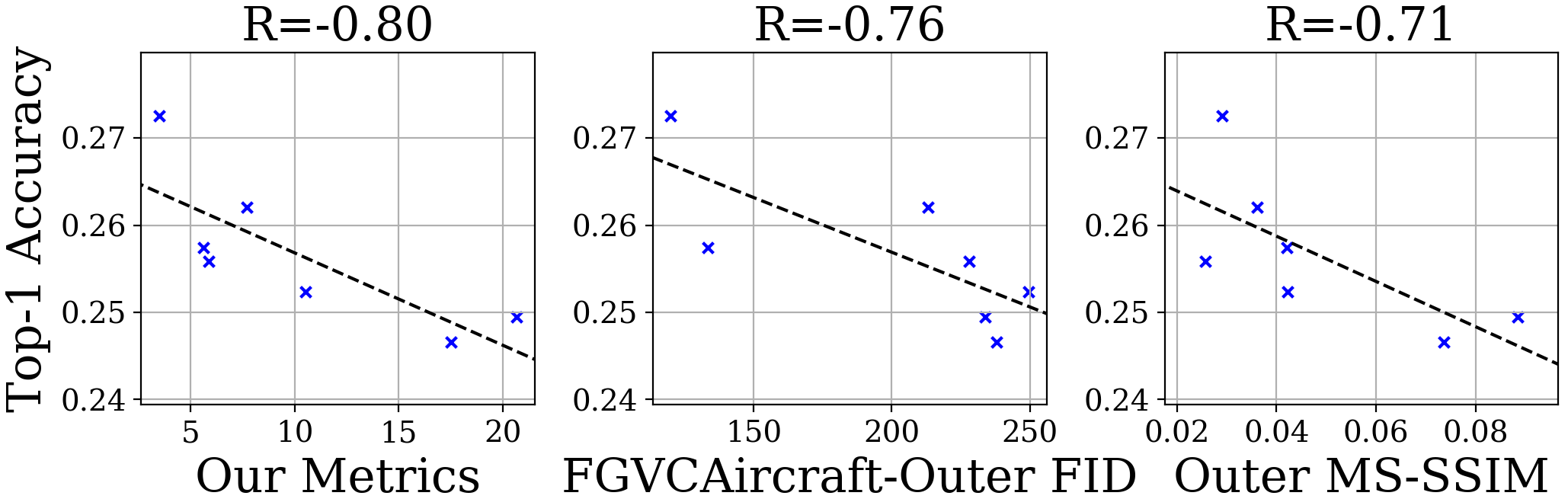}}
        \subfigure[Indoor Scene Recognition]{\label{fig:eval_m_isr}\includegraphics[width=8cm]{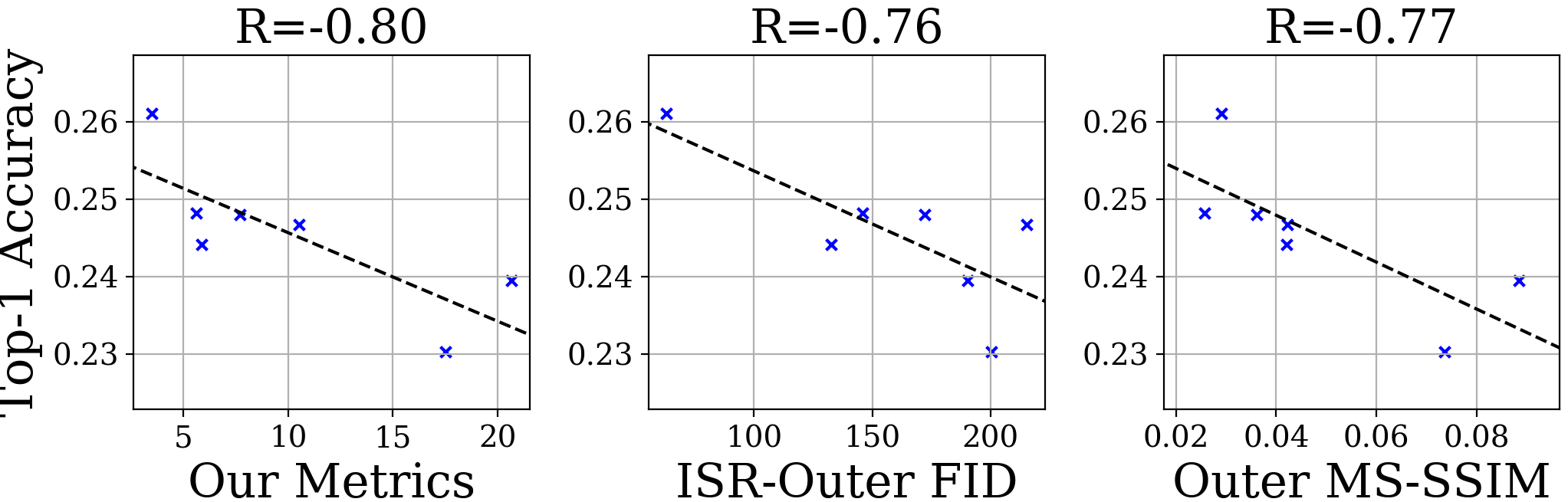}}
        \caption{Comparison of metrics}
        \label{fig:eval_m}
\end{figure}
\begin{figure*}[t]
    \centering
    \includegraphics[width=15cm]{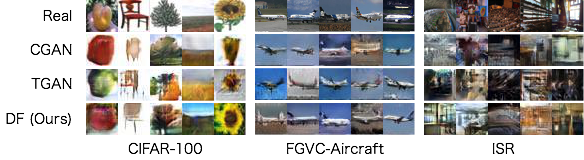}
    \caption{Comparison of generated samples}
    \label{fig:vis_quality}
\end{figure*}

\subsection{Evaluation of Classification Accuracy}\label{sec:eval_cls}
\subsubsection{Comparison to Other GAN-based Data Augmentations}
First, we evaluated the efficacy of \DF~(DF) in terms of the classification accuracy by comparing it to other GAN-based data augmentations.
We compared the performance against two patterns of GAN-based data augmentation: 
generating target samples from (i) CGAN: conditional GANs trained on each target dataset only \cite{Zhu2018bmvc_CGAN_Augmentation}, 
and (ii) TGAN: conditional Transferring GANs pretrained on an outer dataset \cite{wang_transferring_iccv18}.
We also show the performance of classifiers trained on a target dataset without data augmentation (Without DA).

Table~\ref{tb:cmp_gen} lists the results of the top-1 and top-5 accuracy on the classification task, and summarizes the FID and IS of generated samples from GANs.
For the results of DF and TGAN, we report the accuracy with the outer dataset which has best our metric score $\cal{M}$ (Pascal-VOC).
Additionally, for TGAN, we show the best accuracy among 7 outer datasets as TGAN-Best (CIFAR-100 and ISR: Food-101, FGVC-Aircraft: Stanford Cars).
We can see that our DF achieves the best classification accuracy among all patterns.
As reported in~\cite{shmelkov_ECCV18_howgoodismygan}, CGAN dropped the accuracy from Without DA in the cases of CIFAR-100.
On the other hand, we see that DF, which transfers outer knowledge to target models, outperforms Without DA.
DF also generated the target samples with better FID and IS than CGAN.
These results suggest that the quality improvements of the generated samples contribute to the target accuracy.
Compared to TGAN, DF helps more accurate classifications and generates better samples.
For all of the target datasets, we confirmed the differences between DF and TGAN are statistically significant by using the paired t-test with 0.05 of the p-value for all of the top-1/top-5 accuracy, FID, and IS.
These differences may be caused by the transfer strategies of DF and TGAN.
Since TGANs try to transfer outer knowledge by fine-tuning, they suffer from forgetting knowledge~\cite{goodfellow2013empirical} in the pretrained GANs while retraining for the target dataset.
Multi-domain training in DF seems to more effectively transfer the outer knowledge to the target samples without forgetting the knowledge than fine-tuning in TGAN.

In \DF, as shown in Improvements Section, we apply the metric $\cal{M}$ for outer dataset selection and DRS to improve the quality of generated samples and the performance of target classifiers.
As an ablation study, we compare the performances of DF and the cases of DF without our metric $\cal{M}$ and DRS.
Table~\ref{tb:cmp_ablation} shows the results of the ablation study of DF.
Note that the row of DF w/o $\cal{M}$ and DRS denotes the worst cases among outer datasets that use no filtering by DRS, and the outer dataset was LFW for all target datasets.
We see that applying our metric $\cal{M}$ into DF allows us to select an appropriate outer dataset for each target dataset, and DRS boosts the performance of target classifiers and GANs.
Furthermore, we tested CGAN with DRS and TGAN with DRS, but they underperformed our DF in terms of both the accuracy and the sample quality.
This result indicates that DF improves the performances of classifiers and GANs by importing outer dataset knowledge, rather only filtering generated samples by DRS.

\subsubsection{Combining to Conventional Data Augmentation}
We also investigated the classification performance when combining conventional \DA (c\DA) and DF.
For training the classifiers, we adopted multiple \DA transformations: random flip (for x-axis), random expand (100\% to 400\% of expansion ratio), random rotation (0 to 15.0 of angle).
These transformations were applied to images when the images are loaded into a batch.
In Table~\ref{tb:cmp_da}, we show the top-1 and top-5 classification accuracies by applying c\DA and the combination of c\DA and DF.
The outer dataset of DF is Pascal-VOC which has the best our metric score $\cal{M}$ for all target datasets.
In all cases of the target datasets, we see that DF outperforms only using c\DA regarding to the classification performance improvements.
These results indicate that DF generates useful samples that are not obtained from c\DA.

\subsection{Effects of Generated Sample Quality}
The results of Table \ref{tb:cmp_gen} imply that there are a meaningful relation between the target accuracy and the quality of the generated samples.
We analyzed the relation by testing DF on 7 outer datasets.
Figure~\ref{fig:quality_vs_acc} shows the relation between quality (FID and IS) of generated samples from DF on each outer dataset (x-axis) and test accuracy on a target dataset (y-axis).
The dashed line in each panel represents linear regression, and $R$ denotes correlation coefficient.
These plots indicate that the target accuracy depends on the quality of generated samples.
According to these results, DF produces strong or moderate correlations between the test accuracy and both FID and IS.
Further, the visualization results in Figure \ref{fig:vis_quality} show that the samples from DF express more clear features for each class than ones from CGAN and TGAN.
Therefore, we can see that DF improves the target performance because the GANs generates target samples with high quality.

\subsection{Evaluation of Metric ${\cal M}$}\label{sec:eval_metric}
We turn to evaluate our metric ${\cal M}$ for selecting an outer dataset.
We computed ${\cal M}$ by using 5,000 sampled images of each outer dataset and the target datasets.
Figure \ref{fig:eval_m} (left column) represents the relation between our metric ${\cal M}$ and the top-1 accuracy by DF for each outer data.
As the results of the ${\cal M}$ calculation, we obtain ranking of preferable outer datasets for a target dataset.
In this experiment, the ranking order for CIFAR-100 is Pascal-VOC (1.5), Food-101 (2.6), DTD (4.1), Stanford Cars (5.1), Flowers (6.0), SVHN (10.5), LFW (14.2).
For FGVC-Aircraft, the order is Pascal-VOC (3.5), Stanford Cars (5.6), Food-101 (5.9), DTD (7.7), Flowers (10.5), SHVN (17.5), LFW (20.7).
Further, the order of ISR is Pascal-VOC (1.8), Food-101 (3.7), Stanford Cars (5.6), DTD (6.2), Flowers (9.1), SHVN (14.8), LFW (16.8).
By our metric, Pascal-VOC is predicted as the best outer dataset for all of the target datasets.
Since Pascal-VOC is a general image dataset composed of the various modal classes (e.g., Aeroplane, Dogs and Bottles), it has much diversity of the samples ($\overline{\rm SSIM}$ of 0.029).
Moreover, the relevance between each target dataset and Pascal-VOC is also relatively high because Pascal-VOC partially share the classes with the target datasets (CIFAR-100: FID of 50.79, FGVC-Aircraft: FID of 120.05, ISR: FID of 63.2).
From these observations, general datasets such as Pascal-VOC possibly tend to be selected by our metric $\cal{M}$ and to contribute for target models successfully.

The lower score of ${\cal M}$ tends to well predict the higher top-1 accuracy on the classification ($R=-0.99$ in CIFAR-100, $R=-0.80$ in FGVC-Aircraft and ISR).
We also compare ${\cal M}$ to other metrics: FID between the target and each outer dataset, MS-SSIM of the samples of each outer dataset (described in the center and right columns of Figure \ref{fig:eval_m} respectively).
Although the FID and MS-SSIM correlate with the top-1 accuracy, our metric ${\cal M}$ have the equal or stronger correlation than them.
In particular, for FGVC-Aircraft and ISR, our metric ${\cal M}$ succeeds to predict better outer datasets by cooperating FID and MS-SSIM complementarily.

\section{Conclusion}
This paper presented \DF (DF); a generative data augmentation technique based on multi-domain learning GANs.
For improving accuracy in a target task when using a low-volume target dataset, DF exploits outer knowledge via the samples from GANs trained on the target and outer dataset simultaneously.
We also proposed a metric to select the outer dataset that consists of two perspectives: relevance and diversity.
In experiments of the classification task using 3 target and 7 outer datasets, we found that DF improved the target performance and the quality of generated samples.

\clearpage
\appendix
\setcounter{algorithm}{0}
\setcounter{table}{0}
\setcounter{figure}{0}

\section{Appendix}

\begin{algorithm}[h]
    \caption{Modified Discriminator Rejection Sampling for Conditional GANs}
    \label{alg:cdrs} 
    \begin{algorithmic}[1]
        \Require Generator $G$, discriminator $D$ and target label set $Y_{\rm T}$
        \Ensure Filtered class conditional samples from $G$
        \State  $D^{*}$ $\leftarrow$  {\tt KeepTraining}($D,Y_{\rm T}$)
        \For { $y_i$ in $Y_{\rm T}$ }
            \State $\bar{M}_{i}$ $\leftarrow$ {\tt BurnIn}($G, D^{*}, y_i, \tau$)
            \State samples$_i$ $\leftarrow$ $\emptyset$
            \While { $|$samples$_i$$|$ $<$ $N$}
                \State $x_i$ $\leftarrow$ {\tt GetSample}($G,y_i$)
                \State ratio$_i$ $\leftarrow$ $\exp(\tilde{D}^{*}(x_i,y_i))$
                \State $\bar{M}_{i}$ $\leftarrow$ {\tt max}($\bar{M}_i,$ratio$_i$)
                \State acc\_prob$_i$ $\leftarrow$ $\sigma$($\hat{F}$($x_i,\bar{M}_i,\epsilon,r$))
                \State $\psi$ $\leftarrow$ {\tt RandomUniform}($0,1$)
                \If {$\psi$ $\leq$ acc\_prob$_i$ }
                    \State samples$_i$.{\tt append}($x_i$)
                \EndIf
            \EndWhile
        \EndFor
    \end{algorithmic}
\end{algorithm}
\subsection{Modified Conditional DRS Algorithm}
In this section, we describe the detail of modified discriminator rejection sampling (DRS) algorithm for \DF.
The original paper of DRS have shown only the algorithm for unconditional sampling.
For this reason, we cannot directly apply the algorithm because \DF requires conditional sampling for the data augmentation.
Therefore, we modify the DRS algorithm for conditional sampling.

Algorithm \ref{alg:cdrs} represents our modified DRS algorithm for conditional GANs.
The main modification is to compute initial $\bar{M}_{i}$ for each class label $y_i$ in {\tt BurnIn} function (line 3 of Algorithm \ref{alg:cdrs}); the {\tt BurnIn} function performs to find a maximum number of density ratios in constant iterations.
This is because the sampling difficulty is different for each class \cite{brock2018biggan}.
If we set maximum density ratio of the whole class as the initial value, the samples of specific class may lose the diversity.
Additionally, the functions {\tt KeepTraining}, which continues the training of discriminators with early stopping, and {\tt GetSample}, which generates samples from generators, are modified from the original DRS algorithm \cite{azadi_drs_iclr19} to be class-wise.
The other parts of our algorithm work the same as the original DRS.

\begin{table}[t]
    \centering
        \caption{Effects of learning by reverse-side \DF ($\overline{\rm DF}$) from CIFAR-100~(C), FGVC-Aircraft~(F) or ISR~(I) to Pascal-VOC~(P). $\cal{M}$ denotes scores of our proposed metric for outer datasets~(C, F, and I).}
        \label{tb:reverse}
        \scalebox{0.75}{
        \begin{tabular}{lccccc}\toprule
            & Top-1 Acc. & Top-5 Acc. & FID & IS & $\cal{M}$\\
            \midrule
            Without DA & 30.5$\pm$0.8 & 69.7$\pm$0.8 & $-$ & $-$ & $-$ \\
            CGAN & 31.0$\pm$0.9 & 71.6$\pm$1.3 & 75.0$\pm$2.2 & 3.87$\pm$0.14 & $-$ \\
            $\overline{\rm DF}$ (C$+$P) & {\bf 32.6}$\pm${\bf 1.8} & {\bf 72.6}$\pm${\bf 2.3}& {\bf 71.0}$\pm${\bf 0.5} & {\bf 4.05}$\pm${\bf 0.08} & {\bf 1.8} \\
            $\overline{\rm DF}$ (F$+$P) & 29.8$\pm$1.5 & 70.0$\pm$0.8& 81.2$\pm$9.5 & 3.63$\pm$0.04 & 12.6\\
            $\overline{\rm DF}$ (I$+$P) & 30.9$\pm$0.3 & 72.0$\pm$0.2& 72.5$\pm$3.6 & 3.90$\pm$0.13 & 2.0\\
            \bottomrule
        \end{tabular}
        }
\end{table}

\subsection{Reverse-side Evaluation}\label{sec:reverse}
Multi-domain training of DF can be applied bidirectionally, that is, the target and outer dataset are reversible since there is no specialization for fitting the model to only the target distribution.
In this section, we demonstrate the reverse-side evaluation of DF to confirm how the target dataset produces positive or negative effects to outer dataset models.
We call the models flipped the target and outer datasets from ones of DF models as {\em reverse-side DF} ($\overline{\rm DF}$) models.
We tested reverse-side DF by using Pascal-VOC as the target because Pascal-VOC is the best outer dataset for all of the target datasets in the experiments of the main paper.
Conversely, we used CIFAR-100, FGVC-Aircraft and ISR as outer datasets.
In the experiments, all settings with respect to training GANs and classifiers were taken over from Experimental Results Section of the main paper.
The reverse-side DF models are applied DRS when generating target samples as well as normal DF.
We tested the trained target models on the validation set of Pascal-VOC (5,823 images).

Table~\ref{tb:reverse} summarizes the performance results of reverse-side \DF.
We can see that $\overline{\rm DF}$ models from CIFAR-100 to Pascal-VOC (C$+$P) and models from ISR to Pascal-VOC (I$+$P) outperform Without DA and CGAN in terms of the classification accuracy and the generated sample quality as well as the models of normal-directions.
Nevertheless, the generated samples from the $\overline{\rm DF}$ models from FGVC-Aircraft to Pascal-VOC (F$+$P) are inferior in quality and cause negative transfer into the target classifier.
This is because FGVC-Aircraft is not a better outer dataset for Pascal-VOC (${\cal M}=12.6$) than CIFAR-100 (${\cal M}=1.8$) nor ISR (${\cal M}=2.0$) since FGVC-Aircraft has large MS-SSIM; meaning less diversity.
In fact, our metric ${\cal M}$ is asymmetric because MS-SSIM depends on only outer datasets.
Therefore, the reverse-side DF can bring negative effects even if the normal-side DF brings quite improvements.
Thus, we should evaluate whether the combination of a target and an outer is appropriate by calculating our metric ${\cal M}$ for each target, not inferring it from the relations of another direction or other combinations of datasets.

\subsection{Forgetting Knowledge on TGAN}
In the main paper, we discussed that TGAN can inferior to DF with respect to the target performance and the sample quality because of forgetting outer knowledge on GANs.
In this section, we investigate the sample quality of TGANs in outer domain to show that TGANs forget the outer knowledge through the fine-tuning process.
We tested the samples of outer dataset domain generated from TGANs fine-tuned with target dataset as following steps;
(i) we trained CGANs with an outer dataset (for 50,000 iterations), (ii) fine-tuned the CGANs into a target dataset (for 50,000 iterations), (iii) finally we re-fine-tuned the CGANs into the outer dataset (for 1,000 iterations).
Note that since the steps of (i) and (ii) are shared with the experiments on the main paper, we reused the trained models.
We used Pascal-VOC as the outer dataset and used CIFAR-100, FGVC-Aircraft, and ISR as target datasets.

Table~\ref{tb:forget} shows the sample quality (FID and IS) of CGAN(P) and the re-fine-tuned TGANs; CGAN(P) learns only Pascal-VOC.
For comparing \DF with TGAN, we also reprint the results of the reverse-side \DF models shown in Table~\ref{tb:reverse}.
We can see that all TGAN models drop their sampling performances for the outer domain than CGAN(P) which is trained with the outer dataset originally.
This indicates that the TGAN models forget their knowledge about the outer domain, which has been obtained in the first step of the training, through the fine-tuning processes.
On the other hand, the $\overline{\rm DF}$ models except for the case of F$+$P improved the performances.
As discussed in the previous section, the degradation of the F$+$P model caused by the mismatch between the outer (FGVC-Aircraft) and target (Pascal-VOC) datasets.
Therefore, TGANs can forget the outer knowledge and degrade the performance by the fine-tuning, whereas our \DF models keep the knowledge and improve the performance by the multi-domain training of GANs as long as selecting an appropriate outer dataset for the target.

\begin{table}[t]
    \centering
        \caption{Forgetting knowledge on TGAN with respect to the sample quality (FID and IS). Each result was measured by re-fine-tuned TGAN models from Pascal-VOC~(P) to CIFAR-100~(C), FGVC-Aircraft~(F), or ISR~(I)}
        \label{tb:forget}
        \scalebox{0.75}{
        \begin{tabular}{lcc}\toprule
            & FID & IS \\
            \midrule
            CGAN(P) & 75.0$\pm$2.2 & 3.87$\pm$0.14 \\
            \midrule
            TGAN(P$\rightarrow$C$\rightarrow$P) & 89.3$\pm$13.9 & 3.73$\pm$0.51 \\
            TGAN(P$\rightarrow$F$\rightarrow$P) & 81.4$\pm$4.9 & 3.79$\pm$0.05 \\
            TGAN(P$\rightarrow$I$\rightarrow$P) & 87.1$\pm$10.9 & 3.62$\pm$0.34 \\
            \midrule
            $\overline{\rm DF}$ (C$+$P) & 71.0$\pm$0.5 & 4.05$\pm$0.08 \\
            $\overline{\rm DF}$ (F$+$P) & 81.2$\pm$9.5 & 3.63$\pm$0.04 \\
            $\overline{\rm DF}$ (I$+$P) & 72.5$\pm$3.6 & 3.90$\pm$0.13 \\
            \bottomrule
        \end{tabular}
        }
\end{table}

\subsection{Visualization Study}
\begin{figure*}[p]
    \centering
        \subfigure[Real]{\includegraphics[width=8cm]{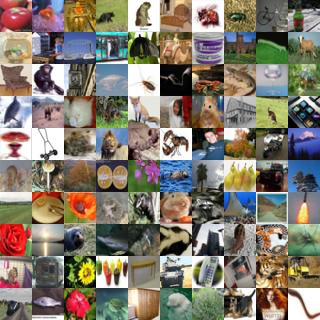}}
        \hspace{3pt}
        \subfigure[CGAN]{\includegraphics[width=8cm]{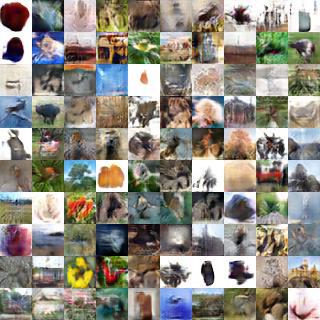}}
        \hspace{3pt}
        \subfigure[TGAN]{\includegraphics[width=8cm]{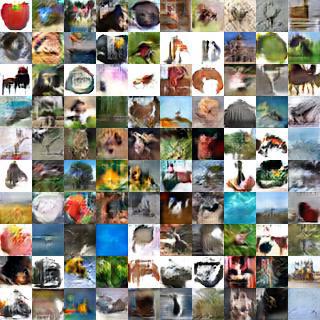}}
        \hspace{3pt}
        \subfigure[\DF]{\includegraphics[width=8cm]{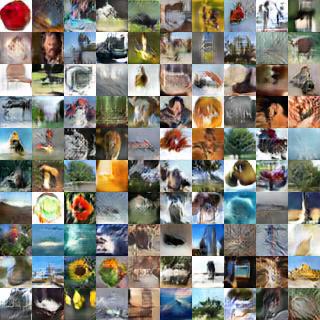}}
        \caption{Comparison of CIFAR-100 real images and generated samples with CGAN, TGAN and \DF}
        \label{fig:vis_cifar}
\end{figure*}

\begin{figure*}[p]
    \centering
        \subfigure[Real]{\includegraphics[width=8cm]{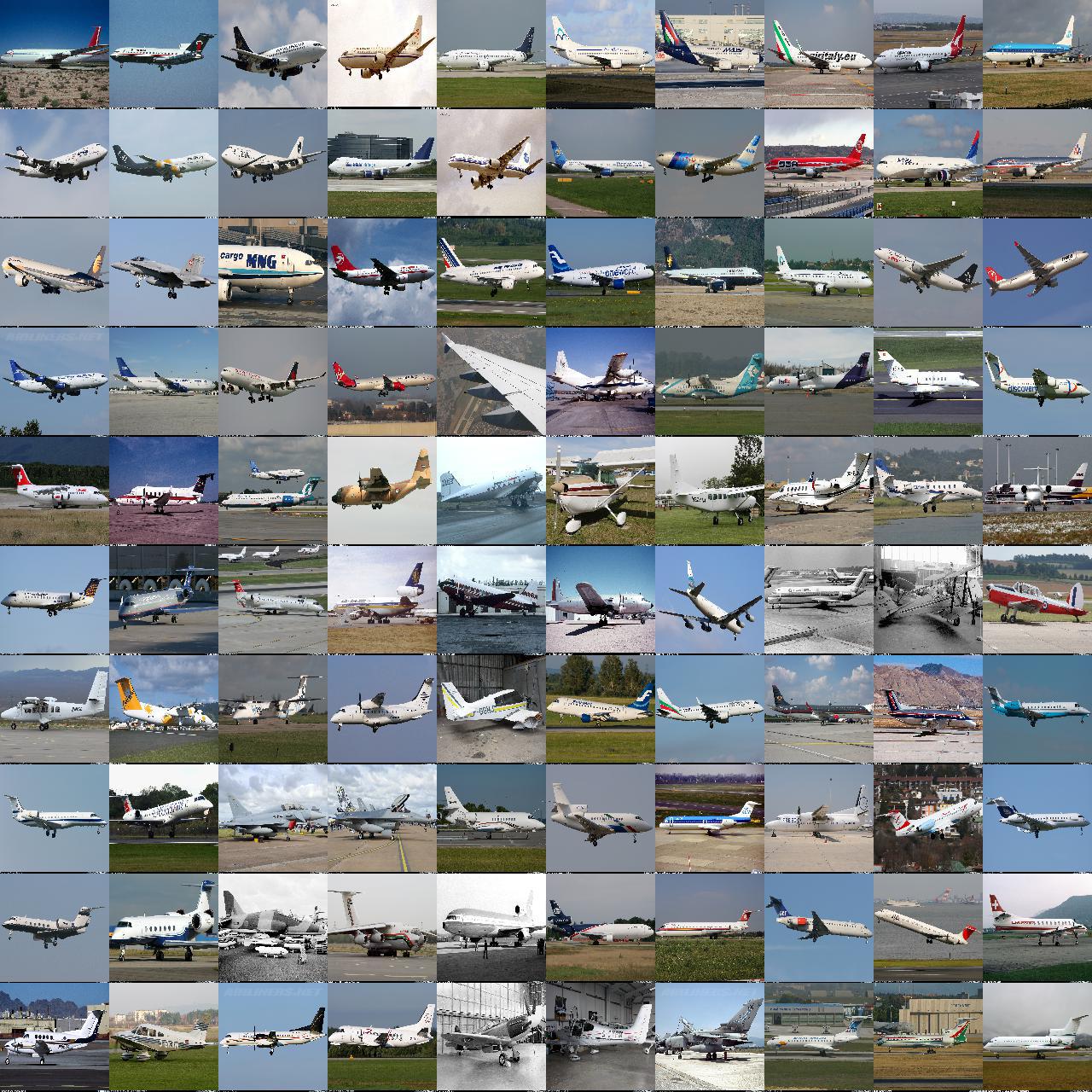}}
        \hspace{3pt}
        \subfigure[CGAN]{\includegraphics[width=8cm]{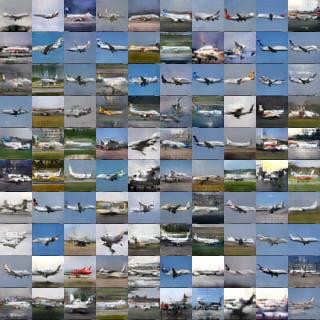}}
        \hspace{3pt}
        \subfigure[TGAN]{\includegraphics[width=8cm]{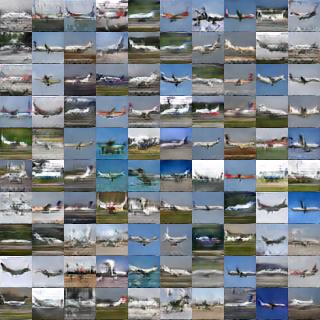}}
        \hspace{3pt}
        \subfigure[\DF]{\includegraphics[width=8cm]{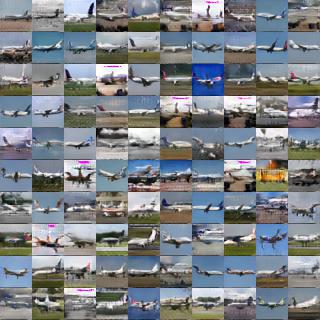}}
        \caption{Comparison of FGVC-Aircraft real images and generated samples with CGAN, TGAN and \DF}
        \label{fig:vis_aircraft}
\end{figure*}

\begin{figure*}[p]
    \centering
        \subfigure[Real]{\includegraphics[width=8cm]{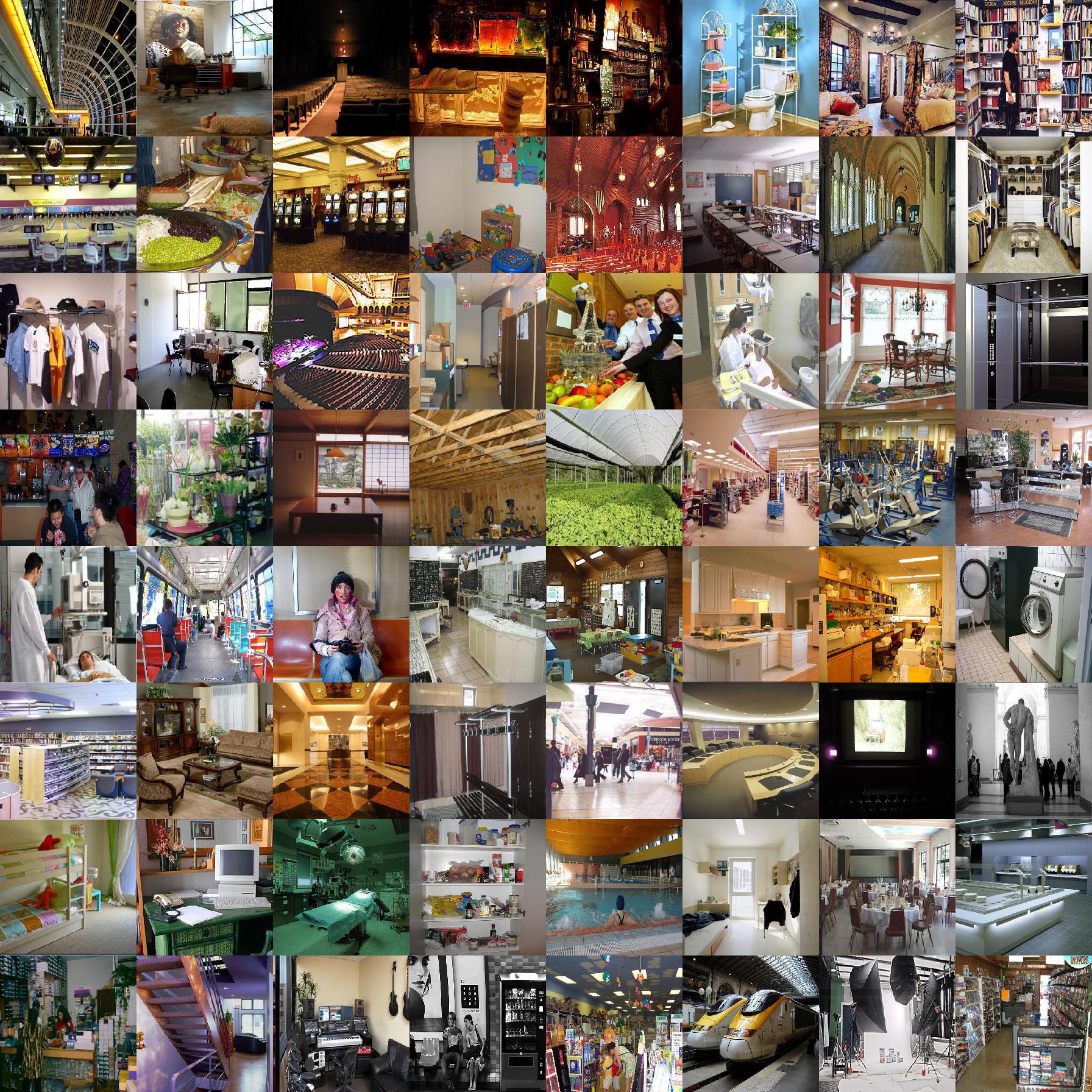}}
        \hspace{3pt}
        \subfigure[CGAN]{\includegraphics[width=8cm]{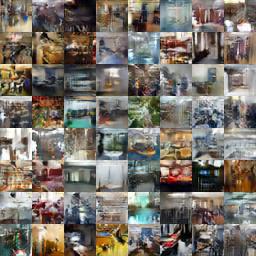}}
        \hspace{3pt}
        \subfigure[TGAN]{\includegraphics[width=8cm]{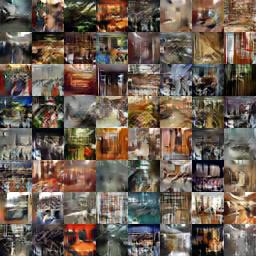}}
        \hspace{3pt}
        \subfigure[\DF]{\includegraphics[width=8cm]{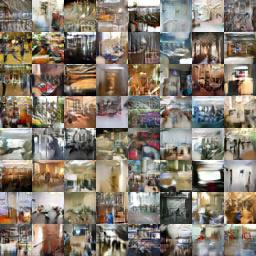}}
        \caption{Comparison of Indoor Scene Recognition real images and generated samples with CGAN, TGAN and \DF}
        \label{fig:vis_isr}
\end{figure*}

\begin{figure}[t]
    \centering
        \subfigure[Real]{\includegraphics[width=4cm]{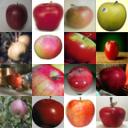}}
        \hspace{2.5pt}
        \subfigure[CGAN]{\includegraphics[width=4cm]{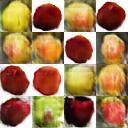}}
        \hspace{2.5pt}
        \subfigure[TGAN]{\includegraphics[width=4cm]{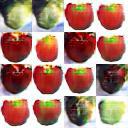}}
        \hspace{2.5pt}
        \subfigure[\DF]{\includegraphics[width=4cm]{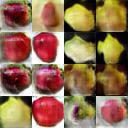}}
        \caption{Comparison of the specific class samples of CIFAR-100 (apples) and the generated images with CGAN, TGAN and \DF}
        \label{fig:vis_cifar_apple}
\end{figure}

\begin{figure}[t]
    \centering
        \subfigure[Real]{\includegraphics[width=4cm]{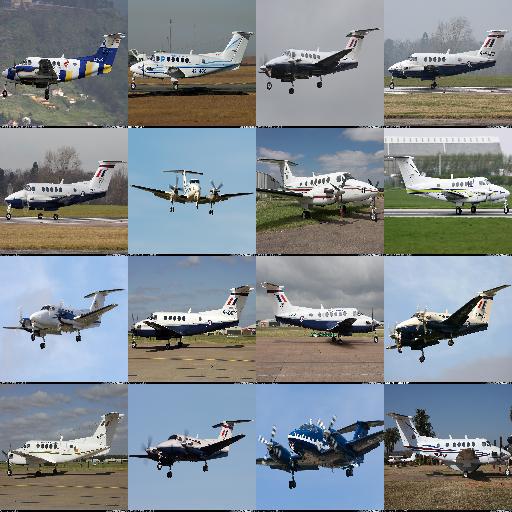}}
        \hspace{2.5pt}
        \subfigure[CGAN]{\includegraphics[width=4cm]{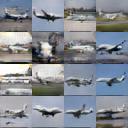}}
        \hspace{2.5pt}
        \subfigure[TGAN]{\includegraphics[width=4cm]{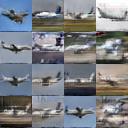}}
        \hspace{2.5pt}
        \subfigure[\DF]{\includegraphics[width=4cm]{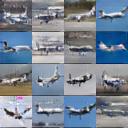}}
        \caption{Comparison of the specific class samples of FGVC-Aircraft (B-200) and the generated images with CGAN, TGAN and \DF}
        \label{fig:vis_aircraft_89}
\end{figure}

\begin{figure}[t]
    \centering
        \subfigure[Real]{\includegraphics[width=4cm]{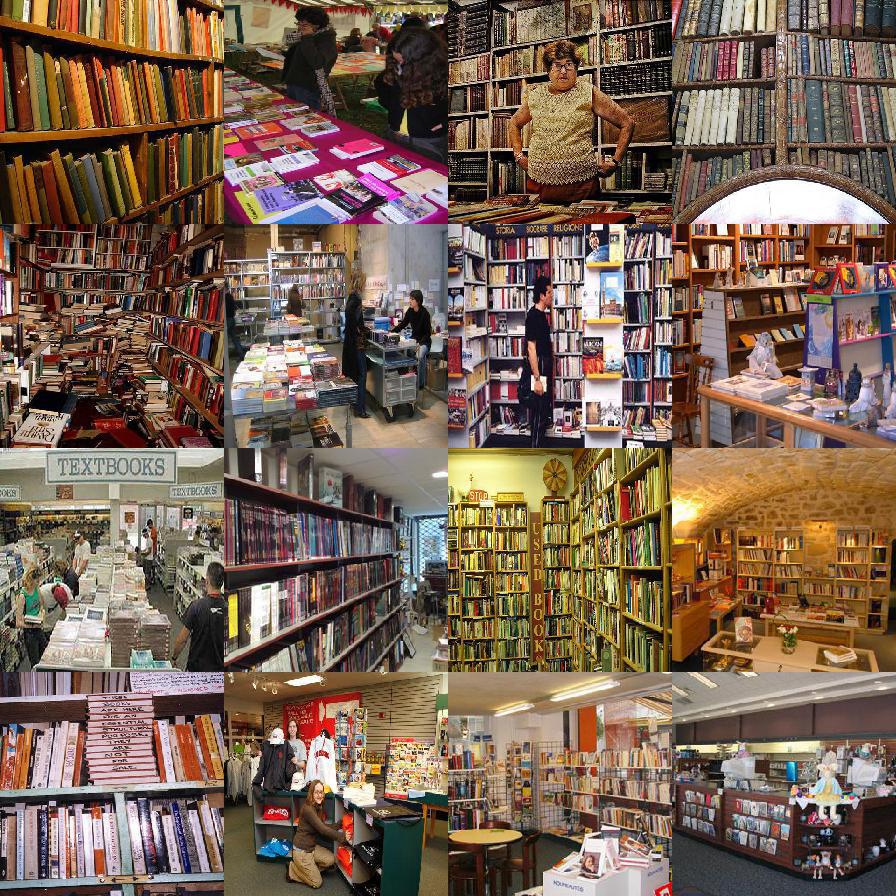}}
        \hspace{2.5pt}
        \subfigure[CGAN]{\includegraphics[width=4cm]{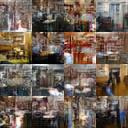}}
        \hspace{2.5pt}
        \subfigure[TGAN]{\includegraphics[width=4cm]{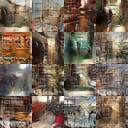}}
        \hspace{2.5pt}
        \subfigure[\DF]{\includegraphics[width=4cm]{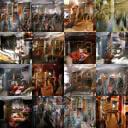}}
        \caption{Comparison of the specific class samples of Indoor Scene Recognition (Bookstore) and the generated images with CGAN, TGAN and \DF}
        \label{fig:vis_isr_7}
\end{figure}

In this section, we demonstrate a qualitative evaluation of generated samples by DF.
Since DF improves target performances and FID/IS, we can expect that DF produces better visual images.
To evaluate the generated images, we compare real images from target dataset to generated images from CGAN, TGAN and DF.
For generating images, we diverted GANs trained in Experimental Results Section of the main paper, i.e. using CIFAR-100, FGVC-Aircraft or ISR as the target datasets, and Pascal-VOC as the outer dataset.
Figure~\ref{fig:vis_cifar}, \ref{fig:vis_aircraft} and \ref{fig:vis_isr} describe images of all classes randomly sampled from real datasets, CGAN, TGAN and DF.
DF generates images with more clear and various shapes than CGAN and TGAN while the difference is slight in visual.

To figure out the detailed characteristics of images from DF, we compare generated images on the specific classes.
Figure~\ref{fig:vis_cifar_apple}, \ref{fig:vis_aircraft_89} and \ref{fig:vis_isr_7} illustrate the images generated from CGAN, TGAN and DF by specifying {\em apples} from CIFAR-100, {\em B-200} from FGVC-Aircraft and {\em Bookstore}, respectively.
In Figure~\ref{fig:vis_cifar_apple}, we can see that DF produces more diverse apple images containing various patterns of pose, background and gloss, whereas images generated by CGAN and TGAN inclined toward few patterns.
For FGVC-Aircraft, Figure~\ref{fig:vis_aircraft_89} shows images of B-200 class, which is characterized by the signature tail unit and the propeller. 
In this case, CGAN and TGAN fails to represents those unique characteristics (e.g., tail units, propellers) on the images, and thus, there is no clear difference from other airplanes.
In contrast, DF can learn fine-grained features of B200 with more fidelity since the characters of tail units and propellers appear on the generated images.
Interestingly, in the case of ISR, the images can contribute to boosting the accuracy of models although the generated images from DF have a slight difference to ones from CGAN and TGAN.
The reason why the little difference in visual for human might be that ISR is originally constructed by the images with higher resolutions than we used in the experiments.
From these visualization studies, we infer that such diversity and fidelity on the images generated by DF helps target models to obtain higher accuracy.

\subsection{Analysis of Classifier}
\begin{table*}[t]
    \begin{center}
        \caption{Top-5 ranking of CIFAR-100 superclass improvement rates by GAN-based data augmentation techniques.}
        \label{tb:rank_superclass}
        \scalebox{0.64}{
        \begin{tabular}{cccccc}\toprule
                 & CGAN                                   & TGAN (Pascal-VOC)                           & DF (Pascal-VOC)                           & DF (Stanford Cars)                   & DF (LFW)\\
            \midrule
            1    & Food Containers (1.18)                 & Food Containers (1.23)                      & Food Containers (1.25)                    & Food Containers (1.24)               & Food Containers (1.13)               \\ 
            2    & Household Furnitures (1.09)            & Household Furnitures (1.22)                 & Household Furnitures (1.21)               & Household Furnitures (1.23)          & Household Furnitures (1.09)          \\ 
            3    & Vehicles2 (1.09)                       & Vehicles2 (1.10)                            & People (1.20)                             & Vehicles1 (1.20)                     & People (1.09)                     \\ 
            4    & Household Electrical Devices (1.00)    & Vehicles1 (1.09)                            & Vehicles2 (1.15)                          & Large Natural Outdoor Scenes (1.13)   & Vehicles2 (1.07)    \\ 
            5    & Large Natural Outdoor Scenes (0.98)     & Large Natural Outdoor Scenes (1.08)          & Vehicles1 (1.12)                          & Vehicles2 (1.11)                     & Large Natural Outdoor Scenes (1.05)                     \\
            \bottomrule
        \end{tabular}
        }
    \end{center}
\end{table*}
As shown in the experiments of the main paper, DF can totally boost accuracy of target classifiers by using knowledge of an outer dataset.
In this section, we further investigate which target class is likely to receive the benefit of DF, and how the difference of outer datasets affects the improvements of DF.
We used the CIFAR-100 classifiers trained in the experiments of Evaluation of Classification Accuracy Section.
For simplicity of the analysis, we tested the classifiers of CIFAR-100 on 20 superclasses~\cite{krizhevsky09_cifar10}, not on the original 100 classes.
Table~\ref{tb:rank_superclass} shows that top-5 ranking of improvements of the classification on superclasses of CIFAR-100 for each GAN-based data augmentation by CGAN, TGAN and DF.
Each ranking was sorted by rates calculated as follows:
\begin{equation}\label{eq:rate}
    \frac{({\rm Accuracy~for~a~superclass~by~a~Without~DA~model})}{({\rm Accuracy~for~a~superclass~by~a~With~DA~model })},
\end{equation}
where a With DA model represents one of CGAN, TGAN, or DF.
Accuracy for each superclass was computed by gathering prediction results for the subclasses, and then calculating top-1 accuracy for the superclass.

In Table~\ref{tb:rank_superclass}, surprisingly, all cases of GAN-based augmentation share the top 1 and 2 highest rated superclasses: Food Containers and Household Furnitures.
This result implies these classes can be easily learned by any GAN-based approaches.
It may be caused by the characteristics of Food Containers and Households Furnitures that tend to be composed of primitive features (e.g., round shapes, lines).
Since these primitive features can be obtained from other classes, GANs can create relatively high quality samples for the two classes.
Comparing DF (Pascal-VOC) with TGAN (Pascal-VOC), DF prominently raises up the performance on People class, which is related to Person class containing in Pascal-VOC.
Contrastively, in TGAN, the improvement of People class is out of ranking.
The improvement rate of TGAN on this class was 1.07 and lower than that of DF (1.20).
This result indicates that DF can effectively transfer the knowledge related to the target tasks from the outer dataset.
Calculating FID/IS scores for People class images, the average scores of DF (FID of 125.6 and IS of 3.85) is superior to the scores of TGAN (FID of 135.5 and IS of 3.29).
This suggests that DF could preserve the knowledge of the outer dataset by multi-domain learning, whereas TGAN may forget it during the process of fine-tuning.

Lastly, we evaluate the effects of outer dataset choices on the classification performance for superclass.
We chose Pascal-VOC, Stanford Cars, and LFW as the outer datasets because they have classes that are similar to the superclasses of the target.
For example, Pascal-VOC has Person class related to People superclass in CIFAR-100 and some classes related to Vehicles1 and Vehicles2 superclass (e.g., Aeroplane, Bus, Car, Train).
In Table~\ref{tb:rank_superclass}, each outer dataset makes a different ranking of the improvement rates except for rank 1 and 2.
We can see that these differences are caused by the classes of the outer datasets since the superclasses appeared in the ranking are related to the class contained by the outer dataset. 
From these results, we conclude that DF enhances target classification performances because it can export the knowledge contained in the outer dataset through the generated samples.

\fontsize{9.0pt}{10.0pt} \selectfont
\bibliographystyle{aaai}
\bibliography{theme}

\end{document}